\pdfoutput=1
\def\year{2021}\relax
\documentclass[letterpaper]{article} 
\usepackage{aaai21}  
\usepackage{times}  
\usepackage{helvet} 
\usepackage{courier}  
\usepackage[hyphens]{url}  
\usepackage{graphicx} 
\usepackage{natbib}  
\usepackage{caption} 
\title{Conversational Neuro-Symbolic Commonsense Reasoning}

\author{Forough Arabshahi\textsuperscript{\rm 1}\footnote{work done when FA and JL were at Carnegie Mellon University.}, Jennifer Lee\textsuperscript{\rm 1}, Mikayla Gawarecki\textsuperscript{\rm 2}\\ Kathryn Mazaitis\textsuperscript{\rm 2}, Amos Azaria\textsuperscript{\rm 3}, Tom Mitchell\textsuperscript{\rm 2}\\ 
}

\affiliations{
\textsuperscript{\rm 1}Facebook,
\textsuperscript{\rm 2}Carnegie Mellon University, \textsuperscript{\rm 3}Ariel University\\
\{forough, jenniferlee98\}@fb.com, \{mgawarec, krivard\}@cs.cmu.edu,\\ amos.azaria@ariel.ac.il, tom.mitchell@cs.cmu.edu}

\usepackage{microtype}
\usepackage{booktabs} 
\usepackage{soul}
\usepackage{epsf,psfrag,amssymb,amsfonts,latexsym,bm,xcolor,array}
\usepackage{subcaption}
\usepackage{caption}
\usepackage{verbatim}
\usepackage[mathscr]{eucal}
\usepackage[tbtags]{mathtools}
\usepackage{multirow}
\usepackage{amsmath}
\usepackage{enumitem}
\usepackage{thmtools,thm-restate}
\usepackage{algorithm}
\usepackage{algpascal}
\usepackage{algc}
\usepackage{algcompatible}
\usepackage[noend]{algpseudocode}
\usepackage[utf8]{inputenc}
\usepackage[english]{babel}
\usepackage{tabularx}
\usepackage{url}
\usepackage{listings}
\usepackage{makecell}
\usepackage{bbold}

\usepackage{tikz}
\usetikzlibrary{
shapes,
shapes.geometric,
shapes.misc,
arrows.meta,
quotes,
graphs,
positioning}

\algnewcommand\algorithmicinput{\textbf{Input:}}
\algnewcommand\Input{\item[\algorithmicinput]}
\algnewcommand\algorithmicoutput{\textbf{Output:}}
\algnewcommand\Output{\item[\algorithmicoutput]}

\definecolor{bondiblue}{rgb}{0.0, 0.58, 0.71}
\definecolor{antiquefuchsia}{rgb}{0.57, 0.36, 0.51}
\definecolor{orange}{rgb}{1,0.5,0}
\definecolor{brickred}{rgb}{0.8, 0.25, 0.33}
\definecolor{gray(x11gray)}{rgb}{0.75, 0.75, 0.75}
\definecolor{lavendergray}{rgb}{0.77, 0.76, 0.82}
\definecolor{lightgray}{rgb}{0.83, 0.83, 0.83}
\definecolor{snow}{rgb}{1.0, 0.98, 0.98}
\definecolor{splashedwhite}{rgb}{1.0, 0.99, 1.0}
\definecolor{timberwolf}{rgb}{0.86, 0.84, 0.82}
\definecolor{seashell}{rgb}{1.0, 0.96, 0.93}
\definecolor{whitesmoke}{rgb}{0.96, 0.96, 0.96}
\definecolor{platinum}{rgb}{0.9, 0.89, 0.89}
\definecolor{pearl}{rgb}{0.94, 0.92, 0.84}
\definecolor{palepink}{rgb}{0.98, 0.85, 0.87}
\definecolor{oldlace}{rgb}{0.99, 0.96, 0.9}
\definecolor{mistyrose}{rgb}{1.0, 0.89, 0.88}
\definecolor{magnolia}{rgb}{0.97, 0.96, 1.0}
\definecolor{lavenderblush}{rgb}{1.0, 0.94, 0.96}
\definecolor{atomictangerine}{rgb}{1.0, 0.6, 0.4}
\definecolor{babyblue}{rgb}{0.54, 0.81, 0.94}
\definecolor{celadon}{rgb}{0.67, 0.88, 0.69}
\definecolor{darkpastelpurple}{rgb}{0.59, 0.44, 0.84}
\definecolor{flamingopink}{rgb}{0.99, 0.56, 0.67}
\definecolor{bluebell}{rgb}{0.64, 0.64, 0.82}
\definecolor{lavenderblue}{rgb}{0.8, 0.8, 1.0}
\definecolor{amethyst}{rgb}{0.6, 0.4, 0.8}
\definecolor{ao(english)}{rgb}{0.0, 0.5, 0.0}
\definecolor{cadmiumorange}{rgb}{0.93, 0.53, 0.18}
\definecolor{darkorange}{rgb}{1.0, 0.55, 0.0}
\definecolor{flame}{rgb}{0.89, 0.35, 0.13}
\definecolor{internationalorange}{rgb}{1.0, 0.31, 0.0}
\definecolor{alizarin}{rgb}{0.82, 0.1, 0.26}
\definecolor{cadmiumred}{rgb}{0.89, 0.0, 0.13}
\sethlcolor{lavenderblush}
\definecolor{candyapplered}{rgb}{1.0, 0.03, 0.0}
\definecolor{carminered}{rgb}{1.0, 0.0, 0.22}
\definecolor{carminepink}{rgb}{0.92, 0.3, 0.26}
\definecolor{coralred}{rgb}{1.0, 0.25, 0.25}

\newcommand{\textGoal}{{\fontfamily{lmtt}\selectfont goal\,}}
\newcommand{\textAction}{{\fontfamily{lmtt}\selectfont action\,}}
\newcommand{\textState}{{\fontfamily{lmtt}\selectfont state\,}}
\newcommand{\art}{\fontfamily{pzc}\selectfont ART}
\newcommand{\CORGI}{CORGI}

\newcommand{\goal}{G(Z)}
\newcommand{\action}{A(Y)}
\newcommand{\state}{S(X)}
\newcommand{\KB}{$\mathcal{K}$}
\newcommand{\blank}{\textcolor{red}{$(\,\cdot^{^{\!\!\mkern-2mu\downarrow}})$}}

\newcommand{\prologTerm}[1]{{\fontfamily{lmss}\selectfont \hl{#1}}}

\newcommand{\orangeTemplate}[2][atomictangerine]{ {\sethlcolor{#1} \hl{#2}} }
\newcommand{\blueTemplate}[2][babyblue]{ {\sethlcolor{#1} \hl{#2}} }
\newcommand{\greenTemplate}[2][celadon]{ {\sethlcolor{#1} \hl{#2}} }

\newcommand{\purpleTemplate}[2][lavenderblue]{ {\sethlcolor{#1} \hl{#2}} }
\newcommand{\redTemplate}[2][coralred]{ {\sethlcolor{#1} \hl{#2}} }

\newcommand{\comet}{$\mathbb{COMET}$}

\newcommand{\largeGap}{\qquad \qquad \qquad \qquad \qquad \qquad \quad}

\begin{document}
\maketitle

\begin{abstract}
In order for conversational AI systems to hold more natural and broad-ranging conversations, they will require much more commonsense, including the ability to identify unstated {\em presumptions} of their conversational partners. For example, in the command ``If it snows at night then wake me up early because I don't want to be late for work'' the speaker relies on commonsense reasoning of the listener to infer the implicit presumption that they wish to be woken only if it snows enough to cause traffic slowdowns.  We consider here the problem of understanding such imprecisely stated natural language commands given in the form of {\bf if-(state), then-(action), because-(goal)} statements.  More precisely, we consider the problem of identifying the unstated {\em presumptions} of the speaker that allow the requested action to achieve the desired goal from the given state (perhaps elaborated by making the implicit presumptions explicit).  We release a benchmark data set for this task, collected from humans and annotated with commonsense presumptions. We present a neuro-symbolic theorem prover that extracts multi-hop reasoning chains, and apply it to this problem.  Furthermore, to accommodate the reality that current AI commonsense systems lack full coverage, we also present an interactive conversational framework built on our neuro-symbolic system, that conversationally evokes commonsense knowledge from humans to complete its reasoning chains.
\end{abstract}

\section{Introduction}
Despite the remarkable success of artificial intelligence (AI) and machine learning in the last few decades, commonsense reasoning remains an unsolved problem at the heart of AI \cite{levesque2012winograd,davis2015commonsense,sakaguchi2019winogrande}. 
Common sense allows us humans to engage in conversations with one another and to convey our thoughts efficiently, without the need to specify much detail \cite{grice1975logic}. For example, if Alice asks Bob to ``wake her up early whenever it snows at night'' so that she can get to work on time, Alice assumes that Bob will wake her up only if it snows enough to cause traffic slowdowns, and only if it is a working day. Alice does not explicitly state these conditions since Bob makes such \emph{presumptions} without much effort thanks to his common sense.
A study, in which we collected such {\bf if(state)-then(action)} commands
from human subjects, revealed that humans often under-specify conditions in their statements; perhaps because they are used to speaking with other humans who possess the common sense needed to infer their more specific intent by making presumptions about their statement.

The inability to make these presumptions makes it challenging 
for computers to engage in natural sounding conversations with humans. While conversational AI systems such as Siri, Alexa, and others are entering our daily lives, their conversations with us humans remains limited to a set of pre-programmed tasks.
We propose that handling unseen tasks requires conversational agents to develop common sense. 

Therefore, we propose a new commonsense reasoning benchmark for conversational agents where the task is to
infer \emph{commonsense presumptions} in commands of the form ``If \textState~holds Then perform \textAction~Because I want to achieve \textGoal.'' The {\bf if-(state), then-(action)} clause arises when humans instruct new conditional tasks to conversational agents \cite{azaria2016instructable, labutov2018lia}. 
The reason for including the \textbf{because-(goal)} clause in the commands is that some presumptions are ambiguous without knowing the user's purpose, or goal. For instance, if Alice's goal in the previous example was to see snow for the first time, Bob would have presumed that even a snow flurry would be excuse enough to wake her up. Since humans frequently omit details when stating such commands, a computer possessing common sense
should be able to infer the hidden \emph{presumptions}; that is, the additional unstated conditions on the If and/or Then portion of the command. Please refer to Tab.~\ref{tab:statement_stats} for some examples. 



\begin{table*}[t]
    \caption{Statistics of \textbf{if-(state), then-(action), because-(goal)} commands collected from a pool of human subjects. The table shows four distinct types of \emph{because}-clauses we found, the count of commands of each type, examples of each and their corresponding commonsense presumption annotations. {\em Restricted domain} includes commands whose \textState is limited to checking email, calendar, maps, alarms, and weather. {\em Everyday domain} includes commands concerning more general day-to-day activities. Annotations are tuples of (index, presumption) where index shows the starting word index of where the missing presumption should be in the command, highlighted with a red arrow. Index starts at 0 and is calculated for the original command.
    }
    \label{tab:statement_stats}
    \centering
    \resizebox{0.85\textwidth}{!}{%
    \begin{tabular}{|cccc|lc|c|}
    \toprule
    \textbf{Domain} & 
        \parbox[b]{3em}{\centering \textbf{\textit{if} clause}} & 
        \parbox[b]{3em}{\centering \textbf{\textit{then} clause}} & 
        \parbox[b]{3em}{\textbf{\textit{because} clause}} & \textbf{Example} &  \parbox[b]{15em}{\centering \textbf{\textit{Annotation:}\\ Commonsense Presumptions}} & \textbf{Count} \\
        \midrule 
         Restricted domain & state & action & goal & \thead[l]{If it's going to rain in the afternoon \blank \\ then remind me to bring an umbrella \blank \\ because I want to remain dry} & \thead{(8, and I am outside) \\ (15, before I leave the house)}  & 76\\
         Restricted domain & state & action & anti-goal & \thead[l]{If I have an upcoming bill payment \blank \\ then remind me to pay it \blank \\ because I don’t want to pay a late fee} & \thead{(7, in the next few days) \\	(13, before the bill payment deadline)} & 3\\
         Restricted domain & state & action & modifier & \thead[l]{If my flight \blank~is from 2am to 4am \\ then book me a supershuttle \blank \\ because it will be difficult to find ubers.} & \thead{(3, take off time) \\ (13, for 2 hours before my flight take off time)} & 2\\
         Restricted domain & state & action & conjunction & \thead[l]{If I receive emails about sales on basketball shoes \blank \\ then let me know \blank \\ because I need them and I want to save money.} & \thead{(9, my size) \\ (13, there is a sale)} & 2\\
         SUM &\ & & & & & 83\\
         \midrule
         Everyday domain & state & action & goal & \thead[l]{If there is an upcoming election  \blank \\ then remind me to register \blank~and vote \blank \\ because I want my voice to be heard.} & \thead{(6, in the next few months) \\ (6, and I am eligible to vote) \\ (11, to vote), (13, in the election)} & 55\\
         Everyday domain & state & action & anti-goal & \thead[l]{If it's been two weeks since my last call with my mentee \\ and I don't have an upcoming appointment with her \blank \\ then remind me to send her an email \blank \\ because we forgot to schedule our next chat} & \thead{(21, in the next few days) \\ (29, to schedule our next appointment)} & 4\\
         Everyday domain & state & action & modifier & \thead[l]{If I have difficulty sleeping \blank \\ then play a lullaby \\ because it soothes me.} & (5, at night) & 12 \\
         Everyday domain & state & action & conjunction &  \thead[l]{If the power goes out \blank \\ then when it comes back on remind me to restart the house furnace \\ because it doesn't come back on by itself and I want to stay warm} & (5, in the Winter) & 6 \\
          SUM &  &  &  &  &  & 77\\
          \bottomrule 
    \end{tabular}
    }
\end{table*}
In this paper, in addition to the proposal of this novel task and the release of a new dataset to study it, we propose a novel initial approach that infers such missing presumptions, by
extracting 
a chain of reasoning that shows how the commanded \textAction will achieve the desired \textGoal when the \textState holds. Whenever any additional reasoning steps appear in this reasoning chain, 
they are output by our system as assumed implicit presumptions associated with the command. For our reasoning method we propose a neuro-symbolic interactive, conversational approach, in which the computer combines its own commonsense knowledge with conversationally evoked knowledge provided by a human user. The reasoning chain is extracted using our neuro-symbolic theorem prover that learns sub-symbolic representations (embeddings) for logical statements, making it robust to variations of natural language encountered in a conversational interaction setting. 
\paragraph{Contributions}
This paper presents three main contributions. 1) We propose a benchmark task for commonsense reasoning in conversational agents and release a data set containing \textbf{if-(state), then-(action), because-(goal)} commands, annotated with commonsense presumptions.
2) We present CORGI (COmmonsense ReasoninG by Instruction), a system that performs soft logical inference. CORGI uses our proposed neuro-symbolic theorem prover and applies it to extract a multi-hop reasoning chain that reveals commonsense presumptions. 
3) We equip CORGI with a conversational interaction mechanism that enables it to collect just-in-time commonsense knowledge from humans.
Our user-study shows (a) the plausibility of relying on humans to evoke commonsense knowledge and (b) the effectiveness of our theorem prover, enabling us to extract reasoning chains for up to 45\% of the studied tasks\footnote{The code and data are available here: https://github.com/ForoughA/CORGI}.

\subsection{Related Work} 
The literature on commonsense reasoning dates back to the very beginning of the field of AI
\cite{winograd1972understanding,mueller2014commonsense,davis2015commonsense} and is studied in several contexts. 
One aspect focuses on building a large knowledge base (KB) of commonsense facts. 
Projects like CYC \cite{lenat1990cyc}, ConceptNet \cite{liu2004conceptnet,havasi2007conceptnet, speer2017conceptnet} and ATOMIC \cite{sap2018atomic,rashkin2018event2mind} are examples of such KBs (see \cite{davis2015commonsense} for a comprehensive list). Recently, \citet{bosselut2019comet} proposed \comet{}, a neural knowledge graph that generates knowledge tuples by learning on examples of structured knowledge.
These KBs provide background knowledge for tasks that require common sense. 
However, it is known that knowledge bases are incomplete, and most have ambiguities and inconsistencies \cite{davis2015commonsense} that must be clarified
for particular reasoning tasks. 
Therefore, we argue that 
reasoning engines can benefit greatly from a \emph{conversational interaction strategy} to ask humans about their missing or inconsistent knowledge. 
Closest in nature to this proposal is the work by \citet{hixon2015learning} on relation extraction through conversation for question answering and \citet{wu2018learning}'s system that learns to form simple concepts through interactive dialogue with a user. 
The advent of intelligent agents and advancements in natural language processing have given learning from conversational interactions a good momentum in the last few years
\citep{azaria2016instructable,labutov2018lia,srivastava2018teaching,goldwasser2014learning,christmann2019look,guo2018dialog,li2018appinite,li2017programming,li2017sugilite}.

A current challenge in commonsense reasoning is lack of benchmarks \cite{davis2015commonsense}. Benchmark tasks in commonsense reasoning include the Winograd Schema Challenge (WSC) \cite{levesque2012winograd}, its variations \cite{kocijan2020review}, and its recently scaled up counterpart, Winogrande \cite{sakaguchi2019winogrande} 
; ROCStories \cite{mostafazadeh2017lsdsem}, COPA \cite{roemmele2011choice}, Triangle COPA \cite{maslan2015one}, and {\art} \cite{bhagavatula2019abductive}, where the task is to choose a plausible outcome, cause or explanation for an input scenario; and the TimeTravel benchmark \cite{qin2019counterfactual} where the task to revise a story to make
it compatible with a given counterfactual
event. 
Other than TimeTravel, most of these benchmarks have a multiple choice design format. 
However, in the real world the computer is usually not given multiple choice questions.
None of these benchmarks targets the extraction of unspoken details in a natural language statement, which is a challenging task for computers known since the 1970's \cite{grice1975logic}. Note than inferring commonsense presumptions is different from intent understanding \cite{janivcek2010abductive,tur2011spoken} where the goal is to understand the intent of a speaker when they say, e.g., ``pick up the mug''. It is also different from implicature and presupposition \cite{sbisa1999presupposition,simons2013conversational,sakama2016abduction} which are concerned with what can be presupposed or implicated by a text.

CORGI has a neuro-symbolic logic theorem prover. Neuro-symbolic systems are hybrid models that leverage the robustness of connectionist methods and the soundness of symbolic reasoning to effectively integrate learning and reasoning \cite{garcez2015neural,besold2017neural}. They have shown promise in different areas of logical reasoning ranging from classical logic to propositional logic, probabilistic logic, abductive logic, and inductive logic \cite{mao2019neuro, manhaeve2018deepproblog,dong2019neural,marra2019integrating,zhou2019abductive,evans2018learning}. To the best of our knowledge, neuro-symbolic solutions for commonsense reasoning have not been proposed before. Examples of commonsense reasoning engines are: AnalogySpace \cite{speer2008analogyspace,havasi2009digital} that uses dimensionality reduction
and \citet{mueller2014commonsense} that uses the event calculus formal language. 
TensorLog \citep{cohen2016tensorlog} converts a first-order logical database into a factor graph and proposes a differentiable strategy for belief propagation over the graph. DeepProbLog \cite{manhaeve2018deepproblog} developed a probabilistic logic programming language 
that is suitable for applications containing categorical variables. 
Contrary to our approach, both these methods do not learn embeddings for logical rules that are needed
to make CORGI robust to natural language variations. 
Therefore, we propose an end-to-end differentiable solution that uses a Prolog \cite{colmerauer1990introduction} proof trace to learn rule embeddings from data. Our proposal is closest to the neural programmer interpreter \citep{reed2015neural} that uses the trace of algorithms such as addition and sort to learn their execution. 
The use of Prolog for performing multi-hop logical reasoning has been studied in \citet{rocktaschel2017end} and \citet{weber2019nlprolog}.
These methods perform Inductive Logic Programming to learn rules from data, and are not applicable to our problem. 
DeepLogic \cite{cingillioglu2018deeplogic}, \citet{rocktaschel2014low}, and  \citet{wang2016blearning} also learn representations for logical rules using neural networks. 
Very recently, transformers were used for temporal logic \cite{finkbeiner2020teaching} and to do multi-hop reasoning \cite{clark2020transformers} using logical facts and rules stated in natural language. 
A purely connectionist approach to reasoning suffers from some limitations. For example, the input token size limit of transformers restricts \citet{clark2020transformers} to small knowledge bases. Moreover, generalizing to arbitrary number of variables or an arbitrary inference depth is not trivial for them. 
Since symbolic reasoning can inherently handle all these challenges, a hybrid approach to reasoning takes the burden of handling them off of the neural component. 
\section{Proposed Commonsense Reasoning Benchmark}
\label{sec:benchmark}
The benchmark task that we propose in this work is that of uncovering hidden \emph{commonsense presumptions} given commands that follow the general format ``if $\langle$state holds$\rangle$ then $\langle$perform action$\rangle$ because $\langle$I want to achieve goal$\rangle$''. 
We refer to these as if-then-because commands. 
We refer to the \emph{if}-clause 
as the \textState, the \emph{then}-clause as the \textAction and the \emph{because}-clause as the \textGoal. 
These natural language commands were collected from a pool of human subjects (more details in the Appendix). 
The data is annotated with unspoken commonsense presumptions by a team of annotators. Tab.~\ref{tab:statement_stats} shows the statistics of the data and annotated examples from the data.
We collected two sets of if-then-because commands.
The first set contains 83 commands targeted at a \textState that can be observed by a computer/mobile phone (
e.g. checking emails, calendar, maps, alarms, and weather). The second set contains 77 commands whose \textState is about day-to-day events and activities. 81\% of the commands over both sets qualify as ``if $\langle \text{\,\textState} \rangle $ then $\langle \text{\,\textAction} \rangle $ because $\langle \text{\,\textGoal} \rangle $''. The remaining 19\% differ in the categorization of the \textit{because}-clause (see Tab.~\ref{tab:statement_stats}); common alternate clause types included anti-goals (``...because I don't want to be late''), modifications of the state or action (``... because it will be difficult to find an Uber''), or conjunctions including at least one non-goal type. Note that we did not instruct the subjects to give us data from these categories, rather we uncovered them after data collection. 
Also, commonsense benchmarks such as the Winograd Schema Challenge \cite{levesque2012winograd} included a similar number of examples (100) when first introduced \cite{kocijan2020review}.

Lastly, 
the if-then-because commands given by humans can be categorized into several different logic templates. 
The discovered logic templates are given in Table \ref{tab:logic_templates} in the Appendix. Our neuro-symbolic theorem prover uses a general reasoning strategy that can address all reasoning templates. However, in an extended discussion in the Appendix, we explain how a reasoning system, including ours, could potentially benefit from these logic templates.

\vspace{-0.5em}
\begin{figure*}[t!]
\centering
\tikzset{%
process/.style  = {rectangle, minimum width=1cm, minimum height=1cm, align=flush center, draw=black, inner sep=0.2cm},
decision/.style = {diamond, minimum width=1cm, minimum height=1cm, align=flush center, draw=black, inner sep=0cm},
resource/.style = {shape=rounded rectangle, minimum width=1cm, minimum height=1cm, align=flush center, draw=black},
stop/.style     = {rectangle, minimum width=1cm, minimum height=1cm, align=flush center, draw=black, double, thick},
waypoint/.style = {coordinate}
}
\newcommand{\cbox}[2]{\parbox{#1}{\centering
#2}}
\newcommand{\mrule}[1]{\rule[0.8ex]{#1}{0.4pt}}
\resizebox{0.9\textwidth}{!}{%
\begin{tikzpicture}[
thick,
>/.tip=Latex,
loose/.style={inner sep 0.7em}]
\draw
	node at (0,0) [font=\large] (statement) {input: If $\langle$\textState$\rangle$ then $\langle$\textAction$\rangle$ because $\langle$\textGoal$\rangle$}
	
	node [process, below of=statement, node distance=1.5cm, left=0.1cm, fill=gray(x11gray)] (parse) {\cbox{3.5cm}{Parse Statement: \\
	$\begin{array}{lcr} \state \leftarrow \text{\textState} \\ 
	\action \leftarrow \text{\textAction} \\
	\goal \leftarrow \text{\textGoal} \end{array}$}}
	
    node [decision,right of=parse,node distance=4cm, fill=lavendergray] (in_k) {\cbox{1.5cm}{Is $G$ in \KB?}}
    
    node [decision,below of=in_k,node distance=2.1cm,minimum width=1.5cm, fill=lavendergray] (loop) {$i>n$?}
    
    node [process,below of=loop,node distance=2.6cm, fill=magnolia] (startFeedback) {\parbox{3.5cm}{Ask the user for more\\ information $G'(Z)$. \\\mrule{3.5cm} \\ $i=i+1$ \\ $\mathrm{goalStack}.push(\goal)$ \\ $ \goal = G' (Z)$}}
    
        node[waypoint,left of=startFeedback,node distance=2.9cm] (loop_0) {}
        node[waypoint,above of=loop_0,node distance=2.5cm] (loop_1) {}
        node[waypoint,above of=loop_1,node distance=2.2cm, right=1.7cm] (loop_2) {}
    
    node[stop,left of=loop,node distance=1.75cm,fill=pearl] (fail0) {Fail}

    node[decision,right of=in_k,node distance=4.5cm,minimum width=3cm, fill=lavendergray] (empty) {\cbox{1.5cm}{goalStack empty?}}
    
    node[process, right of=empty, node distance=4.5cm, minimum width=2cm, fill=lavenderblush] (prover) {\parbox{3cm}{Neuro-Symbolic \\ Theorem Prover: \\ Prove $\goal$}}
    
    node[process,below of=empty,node distance=3cm,label=below:\emph{knowledge base update loop}, fill=magnolia] (add) {\parbox{4cm}{Add a new rule to \KB \\ \mrule{3cm} \\ $\mathrm{goalStack}.top() \vdash \goal$ \\ $\goal = \mathrm{goalStack}.pop()$}}
    
        node[waypoint,left of=add,node distance=2.7cm] (empty_0) {}
        node[waypoint,above of=empty_0,node distance=3cm] (empty_1) {}
    
    node[decision,right of=prover,node distance=4cm,minimum width=3.5cm, fill=lavendergray] (prove) {\cbox{1.5cm}{Is there a proof for $\goal$?}}
    
    node[waypoint, below of=prove, node distance=3cm] (empty_2) {}
    
    node[process, below of=prove, right=0.6cm, node distance=3cm, fill=magnolia] (discard) {\parbox{3.5cm}{discard the rules added in the \emph{knowledge base update loop}}}
    
    node[resource,below of=prover,node distance=2cm, fill=lavenderblush] (embeddings0) {\cbox{3cm}{\textcolor{black}{Rule and Variable embeddings}}}
    
    node[stop,below of=discard,node distance=2cm, fill=pearl] (fail1) {Fail}

    node[decision,right of=prove,node distance=5cm,minimum width=5cm,inner sep=-0.2cm, fill=lavendergray] (check_proof) {\cbox{2.2cm}{Does the \\
    proof contain \\
    $\state$ and $\action$? }}
    
    node[waypoint, below of=check_proof, node distance=3cm] (empty_3) {}
    
    
    node[stop,right of=check_proof,node distance=4cm,label=below:\emph{proof},fill=pearl] (success) {Succeed};
    
\draw(statement) edge [->] (parse);
\draw(parse) edge [->] (in_k);
\draw(in_k) edge ["Y"',pos=0.05,->] (empty);
\draw(in_k) edge ["N",very near start,->] (loop);
\draw(loop) edge ["Y"',->] (fail0);
\draw(loop) edge ["N",very near start,->] (startFeedback);

    \draw(startFeedback) edge (loop_0);
    \draw(loop_0) edge ["\emph{user feedback loop}"] (loop_1);
    \draw(loop_1) edge [->] (loop_2);
\draw(empty) edge ["Y"',->] (prover);
\draw(empty) edge ["N",near start,->] (add);
\draw(prover) edge [->] (prove);

    \draw(add) edge (empty_0);
    \draw(empty_0) edge [->] (empty_1);
\draw(prove) edge ["Y"',->] (check_proof);
\draw(prove) edge ["N",very near start,-] (empty_2);
\draw(empty_2) edge [->] (discard);
\draw(discard) edge [->] (fail1);
\draw(prover) edge [dotted] (embeddings0);
\draw(check_proof) edge ["Y"',->] (success);
\draw(check_proof) edge ["N",near start,-]  (empty_3);
\draw(empty_3) edge [->] (discard);
\end{tikzpicture}
    }%
    
    \caption{CORGI's flowchart. The input is an if-then-because command e.g., ``if it snows tonight then wake me up early because I want to get to work on time''. The input is parsed into its logical form representation (for this example, $\state$ = \prologTerm{weather(snow, Precipitation)}). If CORGI succeeds, it outputs a proof tree for the because-clause or \textGoal (parsed into $\goal$=\prologTerm{get(i,work,on$\_$time)}). The output proof tree contains commonsense presumptions for the input statement (Fig \ref{fig:prooftree} shows an example). If the predicate $G$ does not exist in the knowledge base, \KB, (Is $G$ in \KB?), we have missing knowledge and cannot find a proof. Therefore, we extract it from a human in the \emph{user feedback loop}. At the heart of CORGI is a neuro-symbolic theorem prover that learns rule and variable embeddings to perform a proof (Alg.\ref{alg:inference}). $\mathrm{goalStack}$ and the loop variable $i$ are initialized to empty and $0$ respectively, and $n=3$. \emph{italic text} in the figure represents descriptions that are referred to in the main text. }
    \label{fig:model}
    \vspace{-1.5em}
\end{figure*}
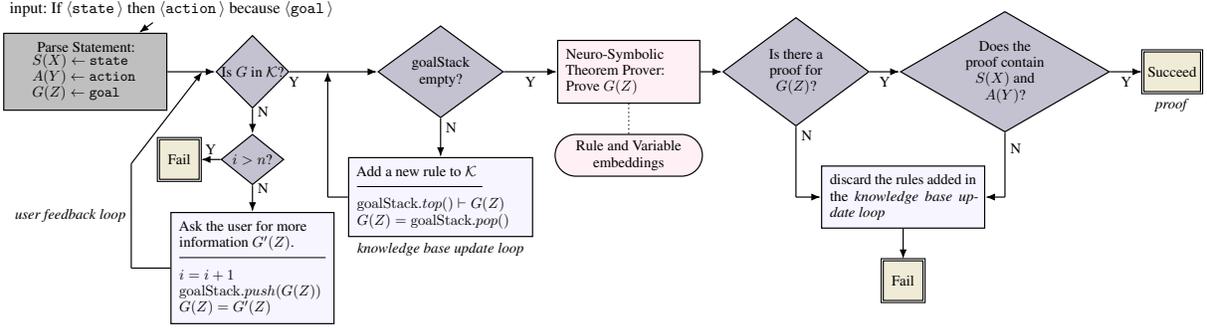
\section{Method}
\label{sec:method}
\paragraph{Background and notation}
The system's commonsense knowledge is a KB, denoted \KB, programmed in a Prolog-like syntax.
We have developed a modified version of Prolog, which has been augmented to support several special features (types, soft-matched predicates and atoms, etc). 
Prolog \cite{colmerauer1990introduction} is a declarative logic programming language that consists of a set of predicates whose arguments are atoms, variables or predicates. A predicate is defined by a set of rules (\prologTerm{$\text{Head} \coloneq \text{Body}.$}) and facts (\prologTerm{$\text{Head}.$}), where \prologTerm{Head} is a predicate, \prologTerm{Body} is a conjuction of predicates, and \prologTerm{$\coloneq$} is logical implication. We use the notation $\state$, $\action$ and $\goal$ to represent the logical form of the \textState, \textAction and \textGoal, respectively where $S$, $A$ and $G$ are predicate names and $X, Y$ and $Z$ indicate the \emph{list} of arguments of each predicate. For example, for \textGoal=``I want to get to work on time'', we have $\goal=$\prologTerm{get(i, work, on\_time)}. Prolog can be used to logically ``prove'' a query (e.g., to prove $\goal$ from $S(X), G(Z)$ and appropriate commonsense knowledge (see the Appendix - Prolog Background)).
\subsection{CORGI: COmmonsense Reasoning by Instruction}
\label{sec:corgi}
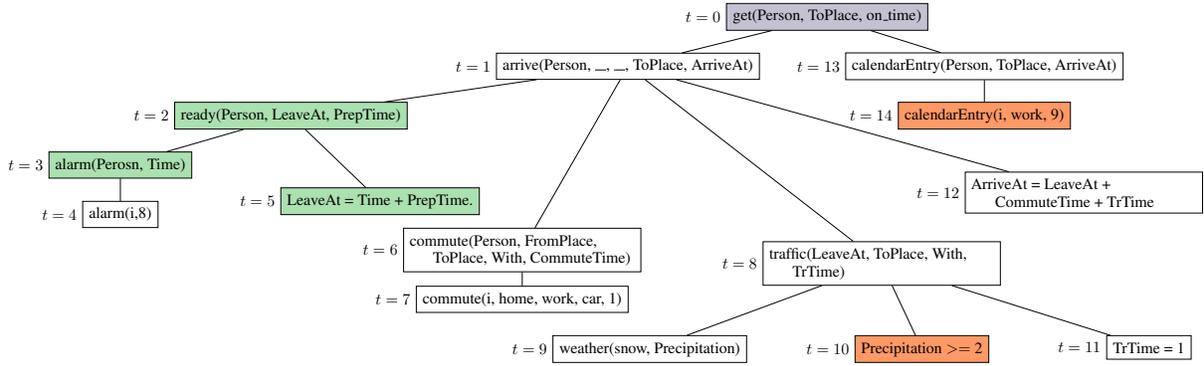
\begin{figure*}\centering
\newcommand{\anonvar}{\rule{0.7em}{0.4pt}}
\newcommand{\rbox}[2]{\parbox{#1}{
\raggedright
\hangindent=1.5em
\hangafter=1
#2}}
\newlength{\pushdown}
\setlength{\pushdown}{0.5cm}

\resizebox{0.9\textwidth}{!}{%
\begin{tikzpicture}
[level distance=1.1cm,
    level 1/.style={sibling distance=7cm},
    level 2/.style={sibling distance=5cm},
    level 3/.style={sibling distance=4cm},
proofstep/.style={rectangle,draw=black,inner sep=4pt}]
\node [proofstep, label=left:{$t=0$}, fill=lavendergray] (get) {get(Person, ToPlace, on$\_$time)}
child { node [proofstep,left=-2cm, label=left:{$t=1$}] (arrive) {arrive(Person, \anonvar, \anonvar, ToPlace, ArriveAt)}
    child { node [proofstep, fill=celadon, label=left:{$t=2$}] (ready) {ready(Person, LeaveAt, PrepTime)}
        child { node [proofstep, fill=celadon, label=left:{$t=3$}, left=0.2cm] (alarm) {alarm(Perosn, Time)} child{node [proofstep, label=left:{$t=4$}] (ground_alarm) {alarm(i,8)} } }
        child { node [proofstep,below left=\pushdown and -2.2cm, fill=celadon, label=left:{$t=5$}] (leaveat) {LeaveAt = Time + PrepTime.}}
    }
    child { node [proofstep,below right=2.5cm and -2.5cm, label=left:{$t=6$}] (commute) {\rbox{5cm}{commute(Person, FromPlace, ToPlace, With, CommuteTime)}}
        child { node [proofstep,below=-0.3cm, label=left:{$t=7$}] (commutex) {commute(i, home, work, car, 1)} }
    }
    child { node [proofstep,below right=2.8cm and 0.5cm, label=left:{$t=8$}] (traffic) {\rbox{5cm}{traffic(LeaveAt, ToPlace, With, TrTime)}} 
        child { node [proofstep,below left=1\pushdown and -1cm, label=left:{$t=9$}] (weather) {weather(snow, Precipitation)} }
        child { node [proofstep,below right=1\pushdown and -0.6cm, fill=atomictangerine, label=left:{$t=10$}] (precipitation) {Precipitation $>=$ 2} }
        child { node [proofstep,below right=1\pushdown and 1cm, label=left:{$t=11$}] (travelTime) {TrTime = 1} }
    }
    child { node [proofstep,below right=2.5\pushdown and 0, label=left:{$t=12$}] (arriveat) {\rbox{5cm}{ArriveAt = LeaveAt + CommuteTime + TrTime}} }
}
child { node [proofstep, label=left:{$t=13$}] (cal) {calendarEntry(Person, ToPlace, ArriveAt)}
    child { node [proofstep, fill=atomictangerine, label=left:{$t=14$}] (calx) {calendarEntry(i, work, 9)} }
};
\end{tikzpicture} 
}
    \caption{Sample proof tree for the \emph{because}-clause of the statement: ``If it snows tonight then wake me up early because I want to get to work on time''. Proof traversal is depth-first from left to right ($t$ gives the order). Each node in the tree indicates a rule's head, and its children indicate the rule's body. For example, the nodes highlighted in green indicate the rule \prologTerm{ready(Person,LeaveAt,PrepTime) $\coloneq$ alarm(Person, Time) $\wedge$ LeaveAt = Time+PrepTime}. The \textGoal we want to prove, $\goal$=\prologTerm{get(Person, ToPlace, on$\_$time)}, is in the tree's root. If a proof is successful, the variables in $\goal$ get grounded (here \prologTerm{Person} and \prologTerm{ToPlace} are grounded to \prologTerm{i} and \prologTerm{work}, respectively). The highlighted orange nodes are the uncovered commonsense presumptions.}
    \label{fig:prooftree}
    \vspace{-1em}
\end{figure*}

CORGI takes as input a natural language command of the form ``if $\langle$\textState$\rangle$ then $\langle$\textAction$\rangle$ because $\langle$\textGoal$\rangle$'' and infers commonsense presumptions by extracting a chain of commonsense knowledge that explains how the commanded \textAction achieves the \textGoal when the \textState holds. For example from a high level, for the command in Fig.~\ref{fig:prooftree} CORGI outputs $(1)$ if it snows more than two inches, then there will be traffic, $(2)$ if there is traffic, then my commute time to work increases, $(3)$ if my commute time to work increases then I need to leave the house earlier to ensure I get to work on time $(4)$ if I wake up earlier then I will leave the house earlier. Formally, this reasoning chain is a proof tree (proof trace) shown in Fig.\ref{fig:prooftree}. 
As shown, the proof tree includes the commonsense presumptions.

CORGI's architecture is depicted in Figure \ref{fig:model}. In the first step, the if-then-because command goes through a parser that extracts the \textState, \textAction and \textGoal from it and converts them to their logical form representations $\state$, $\action$ and $\goal$, respectively.
For example, the \textAction ``wake me up early'' is converted to \prologTerm{wake(me, early)}. The parser is presented in the Appendix (Sec. Parsing). 

The proof trace is obtained by
finding a proof for $\goal$, using 
\KB~and the context of the input if-then-because command. In other words, $\state \cap \action \cap K \rightarrow \goal.$
One challenge is that even the largest knowledge bases gathered to date are incomplete, making it virtually infeasible to prove an arbitrary input $\goal$. Therefore, CORGI is equipped with a conversational interaction strategy, which enables it to prove a query by combining its own commonsense knowledge with conversationally evoked knowledge provided by a human user in response to a question from CORGI (\emph{user feedback loop} in Fig.\ref{fig:model}).
There are 4 possible scenarios that can occur when CORGI asks such questions:
\vspace{-0.5em}
\begin{itemize}
    \item[$\mathfrak{A}$] The user understands the question, but does not know the answer.
    \vspace{-0.4em}
    \item[$\mathfrak{B}$] The user misunderstands the question and responds with an undesired answer.
    \vspace{-0.4em}
    \item[$\mathfrak{C}$] The user understands the question and provides a correct answer, but the system fails to understand the user due to:
    \vspace{-0.4em}
    \begin{itemize}
        \item[$\mathfrak{C}.1$] limitations of natural language understanding.
        \vspace{-0.2em}
        \item[$\mathfrak{C}.2$] variations in natural language, which result in misalignment of the data schema in the knowledge base and the data schema in the user's mind.
    \end{itemize}
    \vspace{-0.5em}
    \item[$\mathfrak{D}$] The user understands the question and provides the correct answer and the system successfully parses and understands it.
\end{itemize}
\vspace{-0.5em}
\CORGI's different components are designed such that they address the above challenges, as explained below.
Since our benchmark data set deals with day-to-day activities, it is unlikely for scenario $\mathfrak{A}$ to occur. If the task required more specific domain knowledge, $\mathfrak{A}$ could have been addressed by choosing a pool of domain experts. Scenario $\mathfrak{B}$ is addressed by asking informative questions from users. 
Scenario $\mathfrak{C}.1$ is addressed by trying to extract small chunks of knowledge from the users piece-by-piece. Specifically, the choice of what to ask the user in the \emph{user feedback loop} is deterministically computed from the user's \textGoal. The first step is to ask how to achieve the user’s stated \textGoal, and CORGI expects an answer that gives a sub-\textGoal. In the next step, CORGI asks 
how to achieve the sub-\textGoal the user just mentioned. 
The reason for this piece-by-piece knowledge extraction is to ensure that the language understanding component can correctly parse the user's response. CORGI then adds the extracted knowledge from the user to \KB~in the \emph{knowledge update loop} shown in Fig.\ref{fig:model}. Missing knowledge outside this \textGoal/sub-\textGoal path is not handled, although it is an interesting future direction. Moreover, the model is user specific and the knowledge extracted from different users are not shared among them. Sharing knowledge raises interesting privacy issues and requires handling personalized conflicts and falls out of the scope of our current study.

Scenario $\mathfrak{C}.2$, caused by the variations of natural language, results in semantically similar statements to get mapped into different logical forms, which is unwanted. For example, ``make sure I am awake early morning'' vs. ``wake me up early morning'' will be parsed into different logical forms \prologTerm{awake(i,early$\_$morning)} and \prologTerm{wake(me, early$\_$morning)},
respectively although they are semantically similar. This mismatch prevents a logical proof from succeeding since the proof strategy relies on exact match in the unification operation (see Appendix). This is addressed by our neuro-symbolic theorem prover (Fig.\ref{fig:model}) that learns vector representations (embeddings) for logical rules and variables and uses them to perform a logical proof through soft unification. If the theorem prover can prove the user's \textGoal, $\goal$, CORGI outputs the proof trace (Fig.\ref{fig:prooftree}) returned by its theorem prover and succeeds. In the next section, we explain our theorem prover in detail.
We revisit scenarios $\mathfrak{A}-\mathfrak{D}$ in detail in the discussion section and show real examples from our user study. 
\section{Neuro-Symbolic Theorem Proving}
\label{sec:softProlog}
Our Neuro-Symbolic theorem prover is a neural modification of \emph{backward chaining} and uses the vector similarity between rule and variable embeddings for unification.
In order to learn these embeddings, our theorem prover learns a general proving strategy by training on proof traces of successful proofs.
From a high level, for a given query our model maximizes the probability of choosing the correct rule to pick in each step of the backward chaining algorithm.
This proposal is an adaptation of Reed et al.'s Neural Programmer-Interpreter \cite{reed2015neural} that learns to execute algorithms such as addition and sort, by training on their execution trace.

In what follows, we represent scalars with lowercase letters, vectors with bold lowercase letters and matrices with bold uppercase letters. 
$\mathbf{M}^{\text{rule}} \in \mathbb{R}^{n_1 \times m_1}$ denotes the embedding matrix for the rules and facts, where $n_1$ is the number of rules and facts and $m_1$ is the embedding dimension. 
$\mathbf{M}^{\text{var}} \in \mathbb{R}^{n_2 \times m_2}$ denotes the variable embedding matrix, where $n_2$ is the number of all the atoms and variables in the knowledge base and $m_2$ is the variable embedding dimension. 
Our knowledge base is type coerced, therefore the variable names are associated with their types (e.g., \prologTerm{alarm(Person,Time)})
\paragraph{Learning}
The model's core consists of an LSTM network whose hidden state indicates the next rule in the proof trace and a proof termination probability, given a query as input. The model has a feed forward network that makes variable binding decisions. The model's training is fully supervised by the proof trace of a query given in a depth-first-traversal order from left to right (Fig.~\ref{fig:prooftree}). The trace is sequentially input to the model in the traversal order as explained in what follows. In step $t \in [0,T]$ of the proof, the model's input is $\epsilon^{inp}_t = \big(\mathbf{q}_t, \mathbf{r}_t, (\mathbf{v}^1_t, \dots, \mathbf{v}^\ell_t) \big)$ and $T$ is the total number of proof steps. $\mathbf{q}_t$ is the query's embedding and is computed by feeding the predicate name of the query into a character RNN. $\mathbf{r}_t$ is the concatenated embeddings of the rules in the parent and the left sister nodes in the proof trace, looked up from $\mathbf{M}^{rule}$. For example in Fig.\ref{fig:prooftree}, $\mathbf{q}_3$ represents the node at proof step $t=3$, $\mathbf{r}_3$ represents the rule highlighted in green (parent rule), and $\mathbf{r}_4$ represents the fact \prologTerm{alarm(i, 8)}. 
The reason for including the left sister node in $\mathbf{r}_t$ is that the proof is conducted in a left-to-right depth first order. Therefore, the decision of what next rule to choose in each node is dependent on both the left sisters and the parent (e.g. the parent and the left sisters of the node at step $t=8$ in Fig.~\ref{fig:prooftree} are the rules at nodes $t=1$, $t=2$, and $t=6$, respectively). 
The arguments of the query are presented in $(\mathbf{v}^1_t, \dots, \mathbf{v}^\ell_t)$ where $\ell$ is the arity of the query predicate. For example, $\mathbf{v}_3^1$ in Fig \ref{fig:prooftree} is the embedding of the variable \prologTerm{Person}. Each $\mathbf{v}^i_t$ for $i \in [0,\ell]$, is looked up from the embedding matrix $\mathbf{M}^{var}$. The output of the model in step $t$ is $\epsilon^{out}_t = \big(c_t, \mathbf{r}_{t+1}, (\mathbf{v}^1_{t+1}, \dots, \mathbf{v}^\ell_{t+1}))$ and is computed through the following equations
\begin{align}
    \mathbf{s}_t &= f_{enc}(\mathbf{q}_t,), ~~
    \mathbf{h}_t = f_{lstm}(\mathbf{s}_t, \mathbf{r}_t ,\mathbf{h}_{t-1}), \\
    c_t &= f_{end}(\mathbf{h}_t), ~~
    \mathbf{r}_{t+1} = f_{rule}(\mathbf{h}_t), ~~ 
    \mathbf{v}^i_{t+1} = f_{var}(\mathbf{v}^i_t), \label{eq:ct} 
\end{align}
where $\mathbf{v}^i_{t+1}$ is a probability vector over all the variables and atoms for the $i^{\text{th}}$ argument, $\mathbf{r}_{t+1}$ is a probability vector over all the rules and facts and $c_t$ is a scalar probability of terminating the proof at step t.
$f_{enc}$, $f_{end}$, $f_{rule}$ and $f_{var}$ are feed forward networks with two fully connected layers, and $f_{lstm}$ is an LSTM network. The trainable parameters of the model are the parameters of the feed forward neural networks, the LSTM network, the character RNN that embeds $\mathbf{q}_t$ and the rule and variable embedding matrices $\mathbf{M}^{rule}$ and $\mathbf{M}^{var}$.

Our model is trained end-to-end. In order to train the model parameters and the embeddings, we maximize the log likelihood probability given below
\begin{equation}
    {\bm \theta}^* = argmax~_{{\bm \theta}} \sum_{\epsilon^{out}, \epsilon^{in}} \log(P(\epsilon^{out} \vert \epsilon^{in};{\bm \theta})),
\end{equation}
where the summation is over all the proof traces in the training set and ${\bm \theta}$ is the trainable parameters of the model. We have
\begin{equation}
    \log(P(\epsilon^{out} \vert \epsilon^{in};{\bm \theta})) = \sum_{t=1}^T \log P(\epsilon^{out}_t \vert \epsilon^{in}_1 \dots \epsilon^{in}_{t-1}; {\bm \theta}), \\
\end{equation}
\begin{align}
    \log P(\epsilon^{out}_t \vert \epsilon^{in}_1 \dots \epsilon^{in}_{t-1}; {\bm \theta}) = & \log P(\epsilon^{out}_t \vert \epsilon^{in}_{t-1}; {\bm \theta}) \nonumber \\ 
     = & \log P(c_{t}\vert \mathbf{h}_t) +  \label{eq:probs}  \nonumber \\
     & \log P(\mathbf{r}_{t+1}\vert \mathbf{h}_t) + \nonumber \\
     & \log \sum_{i} P(\mathbf{v}^i_{t+1}\vert \mathbf{v}^i_t).
\end{align} 
Where the probabilities in Equation \eqref{eq:probs} are given in Equations  \eqref{eq:ct}. The inference algorithm for porving is given in the Appendix, section Inference. 
\section{Experiment Design}
The knowledge base, \KB, used for all experiments is a small handcrafted set of commonsense knowledge that reflects the incompleteness of SOTA KBs. See Tab.~\ref{tab:kb_examples} in the Appendix for examples and the code supplement to further explore our KB.
\KB~includes general information about time, restricted-domains such as setting alarms and notifications, emails, and so on, as well as commonsense knowledge about day-to-day activities. \KB~contains a total of 228 facts and rules. Among these, there are 189 everyday-domain and 39 restricted domain facts and rules. We observed that most of the if-then-because commands require everyday-domain knowledge for reasoning, even if they are restricted-domain commands (see Table \ref{tab:dialog} for example). 

Our Neuro-Symbolic theorem prover is trained on proof traces (proof trees similar to Fig.~\ref{fig:prooftree}) collected by proving automatically generated queries to \KB~using sPyrolog\footnote{\url{https://github.com/leonweber/spyrolog}}. 
$\mathbf{M}^{\text{rule}}$ and $\mathbf{M}^{\text{var}}$ are initialized randomly and with GloVe embeddings \cite{pennington2014glove}, respectively, where $m_1=256$ and $m_2=300$. Since \KB~is type-coerced (e.g. \prologTerm{Time}, \prologTerm{Location}, $\dots$), initializing the variables with pre-trained word embeddings helps capture their semantics and improves the performance. 
The neural components of the theorem prover are implemented in PyTorch \cite{paszke2017automatic} and the prover is built on top of sPyrolog.
\vspace{-0.5em}
\subsection{User Study}
\begin{table*}
\begin{minipage}{0.4\textwidth}
    \caption{percentage of successful reasoning tasks for different user types. In no-feedback, user responses are not considered in the proof attempt. in soft unification CORGI uses our proposed neuro-symbolic theorem prover. In the Oracle scenario, the theorem prover has access to oracle embeddings and soft unification is 100\% accurate.}
    \label{tab:user_study}
    \centering
    \resizebox{1\textwidth}{!}{%
    \begin{tabular}{lcc}
    \toprule
    CORGI variations & Novice User & Expert User  \\ \midrule
No-feedback & 0\% & 0\% \\
Soft unification & 15.61\% & 35.00\%  \\
Oracle unification & 21.62\% & 45.71\% \\
\bottomrule
    \end{tabular}
    }
\end{minipage}
\!
\begin{minipage}{0.6\textwidth}
    \caption{\parbox{6.5cm}{Sample dialogs of 2 novice users in our study. CORGI's responses are noted in italics.}}
    \label{tab:dialog}
\centering
    \resizebox{0.95\textwidth}{!}{
    \begin{tabular}{p{1.7\columnwidth}}
    \toprule
    Successful task\\
    \midrule
    If it's going to rain in the afternoon then remind me to bring an umbrella because I want to remain dry.\\
    \hspace{2em}\textit{How do I know if ``I remain dry''?}\\
    If I have my umbrella.\\
    \hspace{2em}\textit{How do I know if ``I have my umbrella''?}\\
    If you remind me to bring an umbrella.\\
    \hspace{2em}\textit{Okay, I will perform ``remind me to bring an umbrella'' in order to achieve ``I remain dry''.}\\
    \\
    \toprule
    Failed task\\
    \midrule
    If it's going to rain in the afternoon then remind me to bring an umbrella because I want to remain dry.\\
    \hspace{2em}\textit{How do I know if ``I remain dry''?}\\
    If I have my umbrella.\\
    \hspace{2em}\textit{How do I know if ``I have my umbrella''?}\\
    If it's in my office.\\
    \hspace{2em}\textit{How do I know if ``it's in my office''?}\\
    ...
    \\
    \bottomrule
    \end{tabular}
    }
\end{minipage}
\end{table*}
In order to assess CORGI's performance, 
we ran a user study. We selected 10 goal-type if-then-because commands from the dataset in Table \ref{tab:statement_stats} and used each as the prompt for a reasoning task.
We had 28 participants
in the study, 4 of which were experts closely familiar with CORGI and its capabilities. The rest were undergraduate and graduate students with the majority being in engineering or computer science fields and some that majored in business administration or psychology. These users had never interacted with CORGI prior to the study (novice users).
Each person was issued the 10 reasoning tasks, taking on average 20 minutes to complete all 10. 

Solving a reasoning task consists of participating in a dialog with CORGI  as the system attempts to complete a proof for the \textGoal of the current task; 
see sample dialogs in Tab.~\ref{tab:dialog}. The task succeeds if CORGI is able to use the answers provided by the participant to construct a reasoning chain (proof) leading from the \textGoal to the \textState and \textAction.
We collected 469 dialogues in our study.

The user study was run with the architecture shown in Fig.~\ref{fig:model}. 
We used the participant responses from the study to run a few more experiments. We (1) Replace our theorem prover with an \emph{oracle} prover that selects the optimal rule at each proof step in Alg.~\ref{alg:inference} and (2) attempt to prove the \textGoal without using any participant responses (\emph{no-feedback}). Tab.~\ref{tab:user_study} shows the success rate in each setting. 




\vspace{-0.5em}
\subsection{Discussion}
In this section, we analyze the results from the study and provide examples of the 4 scenarios in Section \ref{sec:corgi} that we encountered. As hypothesized, scenario $\mathfrak{A}$ 
hardly occurred as the commands are about day-to-day activities that all users are familiar with.
We did encounter scenario $\mathfrak{B}$, however. The study's dialogs show that some users provided means of \emph{sensing} the \textGoal rather than the \emph{cause} of the \textGoal.
For example, for the reasoning task \emph{``If there are thunderstorms in the forecast within a few hours then remind me to close the windows because I want to keep my home dry''}, in response to the system's prompt \emph{``How do I know if `I keep my home dry'?''} a user responded \emph{``if the floor is not wet''} as opposed to an answer such as \emph{``if the windows are closed''}. Moreover, some users did not pay attention to the context of the reasoning task. For example, another user responded to the above prompt (same reasoning task) 
with \emph{``if the temperature is above 80''}! 
Overall, we noticed that CORGI's ability to successfully reason about an if-then-because statement was heavily dependent on whether the user knew how to give the system what it needed, and not necessarily what it asked for; see Table \ref{tab:dialog} for an example. As it can be seen in Table \ref{tab:user_study}, expert users are able to more effectively provide answers that complete CORGI's reasoning chain, likely because they know that regardless of what CORGI asks, the object of the dialog is to connect the because \textGoal back to the knowledge base in some series of if-then rules (\textGoal/sub-\textGoal path in Sec.\ref{sec:corgi}). Therefore, one interesting future direction is to develop a dynamic context-dependent Natural Language Generation method for asking more effective questions.

We would like to emphasize that although it seems to us, humans, that the previous example requires very simple background knowledge that likely exists in SOTA large commonsense knowledge graphs such as ConcepNet\footnote{\url{http://conceptnet.io/}}, ATOMIC\footnote{\url{https://mosaickg.apps.allenai.org/kg_atomic}} or COMET \cite{bosselut2019comet}, this is not the case (verifiable by querying them online). 
For example, for queries such as \emph{``the windows are closed''}, COMET-ConceptNet generative model\footnote{\url{https://mosaickg.apps.allenai.org/comet_conceptnet}} returns knowledge about blocking the sun, and COMET-ATOMIC generative model\footnote{\url{https://mosaickg.apps.allenai.org/comet_atomic}} returns knowledge about keeping the house warm or avoiding to get hot; which while being correct, is not applicable in this context. For \emph{``my home is dry''}, both COMET-ConceptNet and COMET-ATOMIC generative models return knowledge about house cleaning or house comfort. On the other hand, the fact that 40\% of the novice users in our study were able to help CORGI reason about this example with responses such as \emph{``If I close the windows''} to CORGI's prompt, is an interesting result. This tells us that conversational interactions with humans could pave the way for commonsense reasoning and enable computers to extract just-in-time commonsense knowledge, which would likely either not exist in large knowledge bases or be irrelevant in the context of the particular reasoning task.  
Lastly, we re-iterate that as conversational agents (such as Siri and Alexa) enter people's lives, leveraging conversational interactions for learning has become a more realistic opportunity than ever before.

In order to address scenario $\mathfrak{C}.1$, the conversational prompts of CORGI 
ask for specific small pieces of knowledge that can be easily parsed into a predicate and a set of arguments. However, some users in our study tried to provide additional details, which challenged CORGI's natural language understanding. 
For example, for the reasoning task \emph{``If I receive an email about water shut off then remind me about it a day before because I want to make sure I have access to water when I need it.''}, in response to the system's prompt \emph{``How do I know if `I have access to water when I need it.'?''} one user responded \emph{``If I am reminded about a water shut off I can fill bottles''}. This is a successful knowledge transfer. However, the parser expected this to be broken down into two steps. If this user responded to the prompt with \emph{``If I fill bottles''} first, CORGI would have asked \emph{``How do I know if `I fill bottles'?''} and if the user then responded \emph{``if I am reminded about a water shut off''} CORGI would have succeeded. The success from such conversational interactions are not reflected in the overall performance mainly due to the limitations of natural language understanding.

Table \ref{tab:user_study} evaluates the effectiveness of conversational interactions for proving compared to the no-feedback model. The 0\% success rate there reflects the incompleteness of \KB. The improvement in task success rate between the no-feedback case and the other rows indicates that when it is possible for users to contribute useful common-sense 
knowledge to the system, performance improves. The users contributed a total number of 96 rules to our knowledge base, 31 of which were unique rules. 
Scenario $\mathfrak{C}.2$ occurs when there is variation in the user's natural language statement and is addressed with our neuro-symbolic theorem prover. Rows 2-3 in Table \ref{tab:user_study} evaluate our theorem prover (\emph{soft unification}). 
Having access to the optimal rule for unification 
does still better, but the task success rate is not 100\%, mainly due to the limitations of natural language understanding explained earlier.  

\section{Conclusions}
In this paper, we introduced a benchmark task for commonsense reasoning that aims at uncovering unspoken intents that humans can easily uncover in a given statement by making presumptions supported by their common sense. In order to solve this task, we propose
CORGI (COmmon-sense ReasoninG by Instruction),  a neuro-symbolic theorem prover that performs commonsense reasoning by initiating a conversation with a user. CORGI has access to a small knowledge base of commonsense facts and completes it as she interacts with the user. We further conduct a user study that indicates the possibility of using conversational interactions with humans for evoking commonsense knowledge and verifies the effectiveness of our proposed theorem prover.
\section*{Acknowledgements} This work was supported in part by AFOSR under research contract FA9550201.


\begin{thebibliography}{59}
\providecommand{\natexlab}[1]{#1}
\providecommand{\url}[1]{\texttt{#1}}
\providecommand{\urlprefix}{URL }
\expandafter\ifx\csname urlstyle\endcsname\relax
  \providecommand{\doi}[1]{doi:\discretionary{}{}{}#1}\else
  \providecommand{\doi}{doi:\discretionary{}{}{}\begingroup
  \urlstyle{rm}\Url}\fi

\bibitem[{Azaria, Krishnamurthy, and Mitchell(2016)}]{azaria2016instructable}
Azaria, A.; Krishnamurthy, J.; and Mitchell, T.~M. 2016.
\newblock Instructable intelligent personal agent.
\newblock In \emph{Thirtieth AAAI Conference on Artificial Intelligence}.

\bibitem[{Besold et~al.(2017)Besold, Garcez, Bader, Bowman, Domingos, Hitzler,
  K{\"u}hnberger, Lamb, Lowd, Lima et~al.}]{besold2017neural}
Besold, T.~R.; Garcez, A.~d.; Bader, S.; Bowman, H.; Domingos, P.; Hitzler, P.;
  K{\"u}hnberger, K.-U.; Lamb, L.~C.; Lowd, D.; Lima, P. M.~V.; et~al. 2017.
\newblock Neural-symbolic learning and reasoning: A survey and interpretation.
\newblock \emph{arXiv preprint arXiv:1711.03902} .

\bibitem[{Bhagavatula et~al.(2020)Bhagavatula, Bras, Malaviya, Sakaguchi,
  Holtzman, Rashkin, Downey, Yih, and Choi}]{bhagavatula2019abductive}
Bhagavatula, C.; Bras, R.~L.; Malaviya, C.; Sakaguchi, K.; Holtzman, A.;
  Rashkin, H.; Downey, D.; Yih, S. W.-t.; and Choi, Y. 2020.
\newblock Abductive commonsense reasoning.
\newblock In \emph{International Conference on Learning Representations
  (ICLR)}.

\bibitem[{Bosselut et~al.(2019)Bosselut, Rashkin, Sap, Malaviya, Celikyilmaz,
  and Choi}]{bosselut2019comet}
Bosselut, A.; Rashkin, H.; Sap, M.; Malaviya, C.; Celikyilmaz, A.; and Choi, Y.
  2019.
\newblock COMET: Commonsense Transformers for Automatic Knowledge Graph
  Construction.
\newblock \emph{arXiv preprint arXiv:1906.05317} .

\bibitem[{Christmann et~al.(2019)Christmann, Saha~Roy, Abujabal, Singh, and
  Weikum}]{christmann2019look}
Christmann, P.; Saha~Roy, R.; Abujabal, A.; Singh, J.; and Weikum, G. 2019.
\newblock Look before you Hop: Conversational Question Answering over Knowledge
  Graphs Using Judicious Context Expansion.
\newblock In \emph{Proceedings of the 28th ACM International Conference on
  Information and Knowledge Management}, 729--738.

\bibitem[{Cingillioglu and Russo(2018)}]{cingillioglu2018deeplogic}
Cingillioglu, N.; and Russo, A. 2018.
\newblock DeepLogic: Towards End-to-End Differentiable Logical Reasoning.
\newblock \emph{arXiv preprint arXiv:1805.07433} .

\bibitem[{Clark, Tafjord, and Richardson(2020)}]{clark2020transformers}
Clark, P.; Tafjord, O.; and Richardson, K. 2020.
\newblock Transformers as soft reasoners over language.
\newblock \emph{arXiv preprint arXiv:2002.05867} .

\bibitem[{Cohen(2016)}]{cohen2016tensorlog}
Cohen, W.~W. 2016.
\newblock Tensorlog: A differentiable deductive database.
\newblock \emph{arXiv preprint arXiv:1605.06523} .

\bibitem[{Colmerauer(1990)}]{colmerauer1990introduction}
Colmerauer, A. 1990.
\newblock An introduction to Prolog III.
\newblock In \emph{Computational Logic}, 37--79. Springer.

\bibitem[{Dalvi~Mishra, Tandon, and Clark(2017)}]{dalvi2017domain}
Dalvi~Mishra, B.; Tandon, N.; and Clark, P. 2017.
\newblock Domain-targeted, high precision knowledge extraction.
\newblock \emph{Transactions of the Association for Computational Linguistics}
  5: 233--246.

\bibitem[{Davis and Marcus(2015)}]{davis2015commonsense}
Davis, E.; and Marcus, G. 2015.
\newblock Commonsense reasoning and commonsense knowledge in artificial
  intelligence.
\newblock \emph{Communications of the ACM} 58(9): 92--103.

\bibitem[{Dong et~al.(2019)Dong, Mao, Lin, Wang, Li, and Zhou}]{dong2019neural}
Dong, H.; Mao, J.; Lin, T.; Wang, C.; Li, L.; and Zhou, D. 2019.
\newblock Neural logic machines.
\newblock In \emph{International Conference on Learning Representations
  (ICLR)}.

\bibitem[{Evans and Grefenstette(2018)}]{evans2018learning}
Evans, R.; and Grefenstette, E. 2018.
\newblock Learning explanatory rules from noisy data.
\newblock \emph{Journal of Artificial Intelligence Research} 61: 1--64.

\bibitem[{Finkbeiner et~al.(2020)Finkbeiner, Hahn, Rabe, and
  Schmitt}]{finkbeiner2020teaching}
Finkbeiner, B.; Hahn, C.; Rabe, M.~N.; and Schmitt, F. 2020.
\newblock Teaching Temporal Logics to Neural Networks.
\newblock \emph{arXiv preprint arXiv:2003.04218} .

\bibitem[{Garcez et~al.(2015)Garcez, Besold, De~Raedt, F{\"o}ldiak, Hitzler,
  Icard, K{\"u}hnberger, Lamb, Miikkulainen, and Silver}]{garcez2015neural}
Garcez, A.~d.; Besold, T.~R.; De~Raedt, L.; F{\"o}ldiak, P.; Hitzler, P.;
  Icard, T.; K{\"u}hnberger, K.-U.; Lamb, L.~C.; Miikkulainen, R.; and Silver,
  D.~L. 2015.
\newblock Neural-symbolic learning and reasoning: contributions and challenges.
\newblock In \emph{2015 AAAI Spring Symposium Series}.

\bibitem[{Goldwasser and Roth(2014)}]{goldwasser2014learning}
Goldwasser, D.; and Roth, D. 2014.
\newblock Learning from natural instructions.
\newblock \emph{Machine learning} 94(2): 205--232.

\bibitem[{Grice(1975)}]{grice1975logic}
Grice, H.~P. 1975.
\newblock Logic and conversation.
\newblock In \emph{Speech acts}, 41--58. Brill.

\bibitem[{Guo et~al.(2018)Guo, Tang, Duan, Zhou, and Yin}]{guo2018dialog}
Guo, D.; Tang, D.; Duan, N.; Zhou, M.; and Yin, J. 2018.
\newblock Dialog-to-action: Conversational question answering over a
  large-scale knowledge base.
\newblock In \emph{Advances in Neural Information Processing Systems},
  2942--2951.

\bibitem[{Havasi, Speer, and Alonso(2007)}]{havasi2007conceptnet}
Havasi, C.; Speer, R.; and Alonso, J. 2007.
\newblock ConceptNet 3: a flexible, multilingual semantic network for common
  sense knowledge.
\newblock In \emph{Recent advances in natural language processing}, 27--29.
  Citeseer.

\bibitem[{Havasi et~al.(2009)Havasi, Speer, Pustejovsky, and
  Lieberman}]{havasi2009digital}
Havasi, C.; Speer, R.; Pustejovsky, J.; and Lieberman, H. 2009.
\newblock Digital intuition: Applying common sense using dimensionality
  reduction.
\newblock \emph{IEEE Intelligent systems} 24(4): 24--35.

\bibitem[{Hixon, Clark, and Hajishirzi(2015)}]{hixon2015learning}
Hixon, B.; Clark, P.; and Hajishirzi, H. 2015.
\newblock Learning knowledge graphs for question answering through
  conversational dialog.
\newblock In \emph{Proceedings of the 2015 Conference of the North American
  Chapter of the Association for Computational Linguistics: Human Language
  Technologies}, 851--861.

\bibitem[{Honnibal and Montani(2017)}]{honnibal2017spacy}
Honnibal, M.; and Montani, I. 2017.
\newblock spaCy 2: Natural Language Understanding with Bloom Embeddings.
\newblock \emph{Convolutional Neural Networks and Incremental Parsing} .

\bibitem[{Jan{\'\i}{\v{c}}ek(2010)}]{janivcek2010abductive}
Jan{\'\i}{\v{c}}ek, M. 2010.
\newblock Abductive reasoning for continual dialogue understanding.
\newblock In \emph{New Directions in Logic, Language and Computation}, 16--31.
  Springer.

\bibitem[{Kocijan et~al.(2020)Kocijan, Lukasiewicz, Davis, Marcus, and
  Morgenstern}]{kocijan2020review}
Kocijan, V.; Lukasiewicz, T.; Davis, E.; Marcus, G.; and Morgenstern, L. 2020.
\newblock A Review of Winograd Schema Challenge Datasets and Approaches.
\newblock \emph{arXiv preprint arXiv:2004.13831} .

\bibitem[{Labutov, Srivastava, and Mitchell(2018)}]{labutov2018lia}
Labutov, I.; Srivastava, S.; and Mitchell, T. 2018.
\newblock LIA: A natural language programmable personal assistant.
\newblock In \emph{Proceedings of the 2018 Conference on Empirical Methods in
  Natural Language Processing: System Demonstrations}, 145--150.

\bibitem[{Lenat et~al.(1990)Lenat, Guha, Pittman, Pratt, and
  Shepherd}]{lenat1990cyc}
Lenat, D.~B.; Guha, R.~V.; Pittman, K.; Pratt, D.; and Shepherd, M. 1990.
\newblock Cyc: toward programs with common sense.
\newblock \emph{Communications of the ACM} 33(8): 30--49.

\bibitem[{Levesque, Davis, and Morgenstern(2012)}]{levesque2012winograd}
Levesque, H.; Davis, E.; and Morgenstern, L. 2012.
\newblock The winograd schema challenge.
\newblock In \emph{Thirteenth International Conference on the Principles of
  Knowledge Representation and Reasoning}.

\bibitem[{Li, Azaria, and Myers(2017)}]{li2017sugilite}
Li, T. J.-J.; Azaria, A.; and Myers, B.~A. 2017.
\newblock SUGILITE: creating multimodal smartphone automation by demonstration.
\newblock In \emph{Proceedings of the 2017 CHI Conference on Human Factors in
  Computing Systems}, 6038--6049. ACM.

\bibitem[{Li et~al.(2018)Li, Labutov, Li, Zhang, Shi, Ding, Mitchell, and
  Myers}]{li2018appinite}
Li, T. J.-J.; Labutov, I.; Li, X.~N.; Zhang, X.; Shi, W.; Ding, W.; Mitchell,
  T.~M.; and Myers, B.~A. 2018.
\newblock APPINITE: A Multi-Modal Interface for Specifying Data Descriptions in
  Programming by Demonstration Using Natural Language Instructions.
\newblock In \emph{2018 IEEE Symposium on Visual Languages and Human-Centric
  Computing (VL/HCC)}, 105--114. IEEE.

\bibitem[{Li et~al.(2017)Li, Li, Chen, and Myers}]{li2017programming}
Li, T. J.-J.; Li, Y.; Chen, F.; and Myers, B.~A. 2017.
\newblock Programming IoT devices by demonstration using mobile apps.
\newblock In \emph{International Symposium on End User Development}, 3--17.
  Springer.

\bibitem[{Liu and Singh(2004)}]{liu2004conceptnet}
Liu, H.; and Singh, P. 2004.
\newblock ConceptNet—a practical commonsense reasoning tool-kit.
\newblock \emph{BT technology journal} 22(4): 211--226.

\bibitem[{Manhaeve et~al.(2018)Manhaeve, Dumancic, Kimmig, Demeester, and
  De~Raedt}]{manhaeve2018deepproblog}
Manhaeve, R.; Dumancic, S.; Kimmig, A.; Demeester, T.; and De~Raedt, L. 2018.
\newblock Deepproblog: Neural probabilistic logic programming.
\newblock In \emph{Advances in Neural Information Processing Systems},
  3749--3759.

\bibitem[{Mao et~al.(2019)Mao, Gan, Kohli, Tenenbaum, and Wu}]{mao2019neuro}
Mao, J.; Gan, C.; Kohli, P.; Tenenbaum, J.~B.; and Wu, J. 2019.
\newblock The neuro-symbolic concept learner: Interpreting scenes, words, and
  sentences from natural supervision .

\bibitem[{Marra et~al.(2019)Marra, Giannini, Diligenti, and
  Gori}]{marra2019integrating}
Marra, G.; Giannini, F.; Diligenti, M.; and Gori, M. 2019.
\newblock Integrating Learning and Reasoning with Deep Logic Models.
\newblock \emph{arXiv preprint arXiv:1901.04195} .

\bibitem[{Maslan, Roemmele, and Gordon(2015)}]{maslan2015one}
Maslan, N.; Roemmele, M.; and Gordon, A.~S. 2015.
\newblock One hundred challenge problems for logical formalizations of
  commonsense psychology.
\newblock In \emph{AAAI Spring Symposium Series}.

\bibitem[{Mostafazadeh et~al.(2017)Mostafazadeh, Roth, Louis, Chambers, and
  Allen}]{mostafazadeh2017lsdsem}
Mostafazadeh, N.; Roth, M.; Louis, A.; Chambers, N.; and Allen, J. 2017.
\newblock Lsdsem 2017 shared task: The story cloze test.
\newblock In \emph{Proceedings of the 2nd Workshop on Linking Models of
  Lexical, Sentential and Discourse-level Semantics}, 46--51.

\bibitem[{Mueller(2014)}]{mueller2014commonsense}
Mueller, E.~T. 2014.
\newblock \emph{Commonsense reasoning: an event calculus based approach}.
\newblock Morgan Kaufmann.

\bibitem[{Paszke et~al.(2017)Paszke, Gross, Chintala, Chanan, Yang, DeVito,
  Lin, Desmaison, Antiga, and Lerer}]{paszke2017automatic}
Paszke, A.; Gross, S.; Chintala, S.; Chanan, G.; Yang, E.; DeVito, Z.; Lin, Z.;
  Desmaison, A.; Antiga, L.; and Lerer, A. 2017.
\newblock Automatic differentiation in PyTorch .

\bibitem[{Pennington, Socher, and Manning(2014)}]{pennington2014glove}
Pennington, J.; Socher, R.; and Manning, C. 2014.
\newblock Glove: Global vectors for word representation.
\newblock In \emph{Proceedings of the 2014 conference on empirical methods in
  natural language processing (EMNLP)}, 1532--1543.

\bibitem[{Qin et~al.(2019)Qin, Bosselut, Holtzman, Bhagavatula, Clark, and
  Choi}]{qin2019counterfactual}
Qin, L.; Bosselut, A.; Holtzman, A.; Bhagavatula, C.; Clark, E.; and Choi, Y.
  2019.
\newblock Counterfactual Story Reasoning and Generation.
\newblock In \emph{Proceedings of the 2019 Conference on Empirical Methods in
  Natural Language Processing and the 9th International Joint Conference on
  Natural Language Processing (EMNLP-IJCNLP)}, 5046--5056.

\bibitem[{Rashkin et~al.(2018)Rashkin, Sap, Allaway, Smith, and
  Choi}]{rashkin2018event2mind}
Rashkin, H.; Sap, M.; Allaway, E.; Smith, N.~A.; and Choi, Y. 2018.
\newblock Event2mind: Commonsense inference on events, intents, and reactions.
\newblock \emph{arXiv preprint arXiv:1805.06939} .

\bibitem[{Reed and De~Freitas(2015)}]{reed2015neural}
Reed, S.; and De~Freitas, N. 2015.
\newblock Neural programmer-interpreters.
\newblock \emph{arXiv preprint arXiv:1511.06279} .

\bibitem[{Rockt{\"a}schel et~al.(2014)Rockt{\"a}schel, Bo{\v{s}}njak, Singh,
  and Riedel}]{rocktaschel2014low}
Rockt{\"a}schel, T.; Bo{\v{s}}njak, M.; Singh, S.; and Riedel, S. 2014.
\newblock Low-dimensional embeddings of logic.
\newblock In \emph{Proceedings of the ACL 2014 Workshop on Semantic Parsing},
  45--49.

\bibitem[{Rockt{\"a}schel and Riedel(2017)}]{rocktaschel2017end}
Rockt{\"a}schel, T.; and Riedel, S. 2017.
\newblock End-to-end differentiable proving.
\newblock In \emph{Advances in Neural Information Processing Systems},
  3788--3800.

\bibitem[{Roemmele, Bejan, and Gordon(2011)}]{roemmele2011choice}
Roemmele, M.; Bejan, C.~A.; and Gordon, A.~S. 2011.
\newblock Choice of plausible alternatives: An evaluation of commonsense causal
  reasoning.
\newblock In \emph{2011 AAAI Spring Symposium Series}.

\bibitem[{Sakaguchi et~al.(2019)Sakaguchi, Bras, Bhagavatula, and
  Choi}]{sakaguchi2019winogrande}
Sakaguchi, K.; Bras, R.~L.; Bhagavatula, C.; and Choi, Y. 2019.
\newblock WINOGRANDE: An adversarial winograd schema challenge at scale .

\bibitem[{Sakama and Inoue(2016)}]{sakama2016abduction}
Sakama, C.; and Inoue, K. 2016.
\newblock Abduction, conversational implicature and misleading in human
  dialogues.
\newblock \emph{Logic Journal of the IGPL} 24(4): 526--541.

\bibitem[{Sap et~al.(2019)Sap, Le~Bras, Allaway, Bhagavatula, Lourie, Rashkin,
  Roof, Smith, and Choi}]{sap2018atomic}
Sap, M.; Le~Bras, R.; Allaway, E.; Bhagavatula, C.; Lourie, N.; Rashkin, H.;
  Roof, B.; Smith, N.~A.; and Choi, Y. 2019.
\newblock Atomic: An atlas of machine commonsense for if-then reasoning.
\newblock In \emph{Proceedings of the AAAI Conference on Artificial
  Intelligence}, volume~33, 3027--3035.

\bibitem[{Sbis{\`a}(1999)}]{sbisa1999presupposition}
Sbis{\`a}, M. 1999.
\newblock Presupposition, implicature and context in text understanding.
\newblock In \emph{International and Interdisciplinary Conference on Modeling
  and Using Context}, 324--338. Springer.

\bibitem[{Simons(2013)}]{simons2013conversational}
Simons, M. 2013.
\newblock On the conversational basis of some presuppositions.
\newblock In \emph{Perspectives on linguistic pragmatics}, 329--348. Springer.

\bibitem[{Speer, Chin, and Havasi(2017)}]{speer2017conceptnet}
Speer, R.; Chin, J.; and Havasi, C. 2017.
\newblock ConceptNet 5.5: An Open Multilingual Graph of General Knowledge.
\newblock 4444--4451.
\newblock
  \urlprefix\url{http://aaai.org/ocs/index.php/AAAI/AAAI17/paper/view/14972}.

\bibitem[{Speer, Havasi, and Lieberman(2008)}]{speer2008analogyspace}
Speer, R.; Havasi, C.; and Lieberman, H. 2008.
\newblock AnalogySpace: Reducing the Dimensionality of Common Sense Knowledge.
\newblock In \emph{AAAI}, volume~8, 548--553.

\bibitem[{Srivastava(2018)}]{srivastava2018teaching}
Srivastava, S. 2018.
\newblock \emph{Teaching Machines to Classify from Natural Language
  Interactions}.
\newblock Ph.D. thesis, Samsung Electronics.

\bibitem[{Tur and De~Mori(2011)}]{tur2011spoken}
Tur, G.; and De~Mori, R. 2011.
\newblock \emph{Spoken language understanding: Systems for extracting semantic
  information from speech}.
\newblock John Wiley \& Sons.

\bibitem[{Wang and Cohen(2016)}]{wang2016blearning}
Wang, W.~Y.; and Cohen, W.~W. 2016.
\newblock Learning First-Order Logic Embeddings via Matrix Factorization.
\newblock In \emph{IJCAI}, 2132--2138.

\bibitem[{Weber et~al.(2019)Weber, Minervini, M{\"u}nchmeyer, Leser, and
  Rockt{\"a}schel}]{weber2019nlprolog}
Weber, L.; Minervini, P.; M{\"u}nchmeyer, J.; Leser, U.; and Rockt{\"a}schel,
  T. 2019.
\newblock NLprolog: Reasoning with Weak Unification for Question Answering in
  Natural Language.
\newblock In \emph{Proceedings of the 57th Annual Meeting of the Association
  for Computational Linguistics, ACL 2019, Florence, Italy, Volume 1: Long
  Papers}, volume~57. ACL (Association for Computational Linguistics).

\bibitem[{Winograd(1972)}]{winograd1972understanding}
Winograd, T. 1972.
\newblock Understanding natural language.
\newblock \emph{Cognitive psychology} 3(1): 1--191.

\bibitem[{Wu et~al.(2018)Wu, Russo, Law, and Inoue}]{wu2018learning}
Wu, B.; Russo, A.; Law, M.; and Inoue, K. 2018.
\newblock Learning Commonsense Knowledge Through Interactive Dialogue.
\newblock In \emph{Technical Communications of the 34th International
  Conference on Logic Programming (ICLP 2018)}. Schloss
  Dagstuhl-Leibniz-Zentrum fuer Informatik.

\bibitem[{Zhou(2019)}]{zhou2019abductive}
Zhou, Z.-H. 2019.
\newblock Abductive learning: towards bridging machine learning and logical
  reasoning.
\newblock \emph{Science China Information Sciences} 62(7): 76101.

\end{thebibliography}

\section*{Appendix}
\subsection*{Data Collection}
Data collection was done in two stages. In the first stage, we collected if-then-because commands from humans subjects. In the second stage, a team of annotators annotated the data with commonsense presumptions. Below we explain the details of the data collection and annotation process.

In the data collection stage, we asked a pool of human subjects to write commands that follow the general format: if $\langle$ state holds $\rangle$ then $\langle$ perform action $\rangle$ because $\langle$ i want to achieve goal $\rangle$. The subjects were given the following instructions at the time of data collection:

``
Imagine the two following scenarios:

Scenario 1: Imagine you had a personal assistant that has access to your email, calendar, alarm, weather and navigation apps, what are the tasks you would like the assistant to perform for your day-to-day life? And why?

Scenario 2: Now imagine you have an assistant/friend that can understand anything. What would you like that assistant/friend to do for you?

Our goal is to collect data in the format ``If …. then …. because ….''
''

After the data was collected, a team of annotators annotated the commands with additional presumptions that the human subjects have left unspoken. These presumptions were either in the \emph{if}-clause and/or the \emph{then}-clause and examples of them are shown in Tables \ref{tab:statement_stats} and \ref{tab:data_examples}
\begin{table*}[h]
    \caption{Example if-then-because commands in the data and their annotations. Annotations are tuples of (index, missing text) where index shows the starting word index of where the missing text should be in the command. Index starts at 0 and is calculated for the original utterance.}
    \label{tab:data_examples}
    \centering
    \begin{tabular}{||c|c||}
    \toprule
        Utterance & Annotation \\
        \midrule
         \thead{If the temperature \blank~is above 30 degrees \blank \\ then remind me to put the leftovers from last night into the fridge \\ because I want the leftovers to stay fresh} & \thead{(2, inside) \\ (7, Celsius)} \\ 
         \midrule 
         \thead{If it snows \blank~tonight \blank \\ then wake me up early \\ because I want to arrive to work early} & \thead{(3, more than two inches) \\ (4, and it is a working day)} \\ 
         \midrule
         \thead{If it's going to rain in the afternoon \blank \\ then remind me to bring an umbrella \blank \\ because I want to stay dry} & \thead{(8, when I am outside) \\ (15, before I leave the house)} \\
         \bottomrule
    \end{tabular}
\end{table*}

\subsection*{Logic Templates}
As explained in the main text, we uncovered 5 different logic templates, that reflect humans' reasoning, from the data after data collection. The templates are listed in Table \ref{tab:logic_templates}. In what follows, we will explain each template in detail using the examples of each template listed in Tab.~\ref{tab:logic_templates}.

In the blue template (Template 1), the \textState results in a ``bad state'' that causes the \emph{not} of the goal. The speaker asks for the \textAction in order to avoid the bad state and achieve the \textGoal. For instance, consider the example for the blue template in Table \ref{tab:logic_templates}. The \textState of snowing a lot at night, will result in a bad state of traffic slowdowns which in turn causes the speaker to be late for work. In order to overcome this bad state. The speaker would like to take the \textAction, waking up earlier, to account for the possible slowdowns cause by snow and get to work on time.

In the orange template (Template 2), performing the \textAction when the \textState holds allows the speaker to achieve the \textGoal and not performing the \textAction when the \textState holds prevents the speaker from achieving the \textGoal. For instance, in the example for the orange template in Table \ref{tab:logic_templates} the speaker would like to know who the attendees of a meeting are when the speaker is walking to that meeting so that the speaker is prepared for the meeting and that if the speaker is not reminded of this, he/she will not be able to properly prepare for the meeting. 

In the green template (Template 3), performing the \textAction when the \textState holds allows the speaker to take a hidden \textAction that enables him/her to achieve the desired \textGoal. For example, if the speaker is reminded to buy flower bulbs close to the Fall season, he/she will buy and plant the flowers (hidden \textAction s) that allows the speaker to have a pretty spring garden.

In the purple template (Template 4), the \textGoal that the speaker has stated is actually a goal that they want to \emph{avoid}. In this case, the \textState causes the speaker's \textGoal, but the speaker would like to take the \textAction when the \textState holds to achieve the opposite of the \textGoal. For the example in Tab.~\ref{tab:statement_stats}, if the speaker has a trip coming up and he/she buys perishables the perishables would go bad. In order for this not to happen, the speaker would like to be reminded not to buy perishables to avoid them going bad while he/she is away. 

The rest of the statements are categorized under the ``other'' category. The majority of these statements contain conjunction in their \textState and are a mix of the above templates. 
A reasoning engine could potentially benefit from these logic templates when performing reasoning. We provide more detail about this in the Extended Discussion section in the Appendix. 

\begin{table*}[t]
    \caption{Different reasoning templates of the statements that we uncovered, presumably reflecting how humans logically reason. $\wedge$, $\neg$, $\coloneq$ indicate logical and, negation, and implication, respectively. $\textAction_h$ is an action that is hidden in the main utterance and $\textAction(\textState)$ indicates performing the $\textAction$ when the $\textState$ holds.}
    \label{tab:logic_templates}
\centering
\resizebox{0.9\textwidth}{!}{%
    \begin{tabular}{llc}
        \toprule 
        \bf{Logic template} & \textbf{Example} & \bf{Count} \\
        \midrule
        1. \thead[l]{\blueTemplate{$(\neg(\text{\textGoal}) \coloneq \text{\textState}) \wedge$} \\ \blueTemplate{$(\text{\textGoal} \coloneq \text{\textAction}(\text{\textState}))$}} & \thead[l]{If it snows tonight \\ then wake me up early \\ because I want to arrive to work on time} & 65 \\
        2. \thead[l]{\orangeTemplate{$(\text{\textGoal} \coloneq \text{\textAction}(\text{\textState})) \wedge$} \\ \orangeTemplate{$(\neg(\text{\textGoal}) \coloneq \neg(\text{\textAction}(\text{\textState})))$}} & \thead[l]{If I am walking to a meeting \\ then remind me who else is there \\ because I want to be prepared for the meeting} & 50 \\
        3. \thead[l]{\greenTemplate{$(\text{\textGoal} \coloneq \text{\textAction}_h) \wedge$} \\ \greenTemplate{$(\text{\textAction}_h \coloneq \text{\textAction}(\text{\textState}))$}} & \thead[l]{If we are approaching Fall \\ then remind me to buy flower bulbs \\ because I want to make sure I have a pretty Spring garden.} & 17 \\
        4. \thead[l]{\purpleTemplate{$(\text{\textGoal} \coloneq \text{\textState}) \wedge$} \\  \purpleTemplate{$(\neg(\text{\textGoal}) \coloneq \text{\textAction}(\text{\textState}))$}}
        & \thead[l]{If I am at the grocery store but I have a trip coming up in the next week \\ then remind me not to buy perishables \\ because they will go bad while I am away} & 5 \\
        5. \redTemplate{other} & \thead[l]{If tomorrow is a holiday \\ then ask me if I want to disable or change my alarms \\ because I don't want to wake up early if I don't need to go to work early.}  &23 \\
        \bottomrule
    \end{tabular}
    }
\end{table*}

\subsection*{Prolog Background}
Prolog \cite{colmerauer1990introduction} is a declarative logic programming language. A Prolog program consists of a set of predicates. A predicate has a name (functor) and $N\geq0$ arguments. $N$ is referred to as the arity of the predicate. A predicate with functor name $F$ and arity $N$ is represented as $F(T_1, \dots, T_N)$ where $T_i$'s, for $i\in [1,N]$, are the arguments that are arbitrary Prolog terms. A Prolog term is either an atom, a variable or a compound term (a predicate with arguments). A variable starts with a capital letter (e.g., \prologTerm{Time}) and atoms start with small letters (e.g. \prologTerm{monday}). 
A predicate defines a relationship between its arguments. For example, \prologTerm{isBefore(monday, tuesday)} indicates that the relationship between Monday and Tuesday is that, the former is before the latter.

A predicate is defined by a set of clauses. A clause is either a Prolog~~\emph{fact} or a Prolog \emph{rule}. A Prolog rule is denoted with \prologTerm{$\text{Head} \coloneq \text{Body}.$},
where the \prologTerm{Head} is a predicate, the \prologTerm{Body} is a conjunction (\prologTerm{$\wedge$}) of predicates, \prologTerm{$\coloneq$} is logical implication, and period indicates the end of the clause. The previous rule is an if-then statement that reads ``\emph{if} the Body holds \emph{then} the Head holds''.
A fact is a rule whose body always holds, and is indicated by \prologTerm{Head.} 
, which is equivalent to \prologTerm{Head $\coloneq$  true}. Rows 1-4 in Table \ref{tab:kb_examples} are rules and rows 5-8 are facts.  

Prolog can be used to logically ``prove'' whether a specific query holds or not (For example, to prove that \prologTerm{isAfter(wednesday,thursday)?} is \prologTerm{false} or that \prologTerm{status(i, dry, tuesday)?} is \prologTerm{true} using the Program in Table \ref{tab:kb_examples}). The proof is performed through \emph{backward chaining}, which is a backtracking algorithm that usually employs a depth-first search strategy implemented recursively. 
In each step of the recursion, the input is a query (goal) to prove and the output is the proof's success/failure. in order to prove a query, a rule or fact whose head \emph{unifies} with the query is retrieved from the Prolog program. The proof continues recursively for each predicate in the body of the retrieved rule and succeeds if all the statements in the body of a rule are \prologTerm{true}. The base case (leaf) is when a fact is retrieved from the program.

At the heart of backward chaining is the \emph{unification} operator, which matches the query with a rule's head. Unification first checks if the functor of the query is the same as the functor of the rule head. If they are the same, unification checks the arguments. If the number of arguments or the arity of the predicates do not match unification fails. Otherwise it iterates through the arguments. For each argument pair, if both are grounded atoms unification succeeds if they are exactly the same grounded atoms. If one is a variable and the other is a grounded atom, unification grounds the variable to the atom and succeeds. If both are variables unification succeeds without any variable grounding. The backwards chaining algorithm and the unification operator is depicted in Figure \ref{fig:prooftree_simple}.

\begin{figure*}\centering
\newcommand{\anonvar}{\rule{0.7em}{0.4pt}}
\newcommand{\rbox}[2]{\parbox{#1}{
\raggedright
\hangindent=1.5em
\hangafter=1
#2}}
\setlength{\pushdown}{0.5cm}

\resizebox{0.7\textwidth}{!}{%
\begin{tikzpicture}
[level distance=1.1cm,
    level 1/.style={sibling distance=8cm},
    level 2/.style={sibling distance=6cm},
    level 3/.style={sibling distance=4cm},
proofstep/.style={rectangle,solid,thick,draw=black,inner sep=4pt},
emph/.style={edge from parent/.style={dashed,black,thick,draw}},
norm/.style={edge from parent/.style={solid,black,thick,draw}}]
\node [proofstep] (get) {\textcolor{ao(english)}{status(i, dry, tuesday)}}
child[emph] {node [proofstep] (get_2) {\textcolor{internationalorange}{status(Person1=i, dry, Date1=tuesday)}}
child[norm] {node [proofstep] (commute) { \textcolor{ao(english)}{isInside(Person1=i, Building1, Date1=tuesday)}}
    child[emph] { node [proofstep] (commute_increase) { \textcolor{internationalorange}{isInside(i, home, tuesday)}}
    }
}
child[norm] { node [proofstep] (leave) {\textcolor{ao(english)}{building(Building1)}}
    child[emph] { node [proofstep] (leave_early) { \textcolor{internationalorange}{building(home)}}
    }
  }
};
\end{tikzpicture} 
}
    \caption{Sample simplified proof tree for query \prologTerm{status(i, dry, tuesday)}. dashed edges show successful unification, orange nodes show the head of the rule or fact that is retrieved by the unification operator in each step and green nodes show the query in each proof step. This proof tree is obtained using the Prolog program or \KB~shown in Tab.~\ref{tab:kb_examples}. In the first step, unification goes through all the rules and facts in the table and retrieves rule number 2 whose head unifies with the query. This is because the query and the rule head's functor name is \prologTerm{status} and they both have 3 arguments. Moreover, the arguments all match since \prologTerm{Person1} grounds to atom \prologTerm{i}, grounded atom \prologTerm{dry} matches in both and variable \prologTerm{Date1} grounds to \prologTerm{tuesday}. In the next step, the proof iterates through the predicates in the rule's body, which are \prologTerm{isInside(i, Building1, tuesday)} and \prologTerm{building(Building1)}, to recursively prove them one by one using the same strategy. Each of the predicates in the body become the new query to prove and proof succeeds if all the predicates in the body are proved. Note that once the variables are grounded in the head of the rule they are also grounded in the rule's body.}
    \label{fig:prooftree_simple}
\end{figure*}
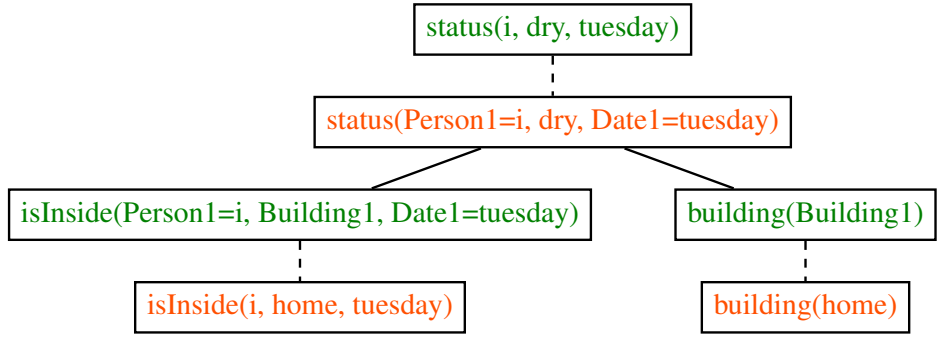
\begin{table}[t]
    \caption{Examples of the commonsense rules and facts in \KB}
    \label{tab:kb_examples}
    \centering
    \resizebox{\columnwidth}{!}{%
    \begin{tabular}{ll}
    \toprule
          1 & \thead[l]{\prologTerm{isEarlierThan(Time1,Time2) :- isBefore(Time1,Time3),} \\ \largeGap ~ \prologTerm{isEarlierThan(Time3,Time2).}} \\
          \\
          2 & \thead[l]{\prologTerm{status(Person1, dry, Date1) :- isInside(Person1, Building1, Date1),} \\ \largeGap \prologTerm{building(Building1).}} \\
          \\
          3 & \thead[l]{ \prologTerm{status(Person1, dry, Date1) :- weatherBad(Date1, \_), } \\ \largeGap \prologTerm{carry(Person1, umbrella, Date1),} \\ \largeGap \prologTerm{isOutside(Person1, Date1).} }\\
          \\
          4 & \prologTerm{notify(Person1, corgi, Action1) :- email(Person1, Action1).} \\
          \\
          5 & \prologTerm{isBefore(monday, tuesday).} \\
          \\
          6 &\prologTerm{has(house, window).} \\
          \\ 
          7 & \prologTerm{isInside(i, home, tuesday).} \\
          \\
          8 & \prologTerm{building(home).} \\
          \bottomrule
    \end{tabular}
    }
\end{table}


\subsection*{Parsing}
The goal of our parser is to extract the \textState, \textAction and \textGoal from the input utterance and convert them to their logical forms $\state$, $\action$, and $\goal$, respectively. The parser is built using Spacy \cite{honnibal2017spacy}. We implement a relation extraction method that uses Spacy's built-in dependency parser. The language model that we used is the en$\_$coref$\_$lg$-3.0.0$ released by Hugging face\footnote{\url{https://github.com/huggingface/neuralcoref-models/releases/download/en_coref_lg-3.0.0/en_coref_lg-3.0.0.tar.gz}}. 
The predicate name is typically the sentence verb or the sentence root. The predicate's arguments are the subject, objects, named entities and noun chunks extracted by Spacy. The output of the relation extractor is matched against the knowledge base through rule-based mechanisms including string matching to decide weather the parsed logical form exists in the knowledge base. 
If a match is found, the parser re-orders the arguments to match the order of the arguments of the predicate retrieved from the knowledge base. This re-ordering is done through a type coercion method. In order to do type coercion, we use the types released by Allen AI in the Aristo tuple KB v1.03 Mar 2017 Release \cite{dalvi2017domain} and have added more entries to it to cover more nouns. 
The released types file is a dictionary that maps different nouns to their types. For example, doctor is of type person and Tuesday is of type date. If no match is found, the parsed predicate will be kept as is and CORGI tries to evoke relevant rules conversationally from humans in the \emph{user feedback loop} in Figure \ref{fig:model}.

We would like to note that we refrained from using a grammar parser, particularly because we want to enable open-domain discussions with the users and save the time required for them to learn the system's language. As a result, the system will learn to adapt to the user's language over time since the background knowledge will be accumulated through user interactions, therefore it will be adapted to that user. A negative effect, however, is that if the parser makes a mistake, error will propagate onto the system's future knowledge. This is an interesting future direction that we are planning to address. 



\subsection*{Inference} The inference algorithm for our proposed neuro-symbolic theorem prover is given in Alg.~\ref{alg:inference}.
In each step $t$ of the proof, given a query $Q$, we calculate $\mathbf{q}_t$
and $\mathbf{r}_t$ from the trained model to compute $\mathbf{r}_{t+1}$. Next, we choose $k$ entries of $\mathbf{M}^{rule}$ corresponding to the top $k$ entries of $\mathbf{r}_{t+1}$ as candidates for the next proof trace. $k$ is set to $5$ and is a tuning parameter. For each rule in the top $k$ rules, we attempt to do variable/argument unification by computing the cosine similarity between the arguments of $Q$ and the arguments of the rule's head. If all the corresponding pair of arguments in $Q$ and the rule's head have a similarity higher than threshold, $T_1=0.9$, unification succeeds, otherwise it fails. 
If unification succeeds, we move to prove the body of that rule. If not, we move to the next rule. 
\begin{algorithm}[t]
\caption{Neuro-Symbolic Theorem Prover}
\label{alg:inference}
\begin{algorithmic}
    \Input \textGoal $\goal$, $\bf{M}^{rule}$, $\bf{M}^{var}$, Model parameters, threshold $T_1$, $T_2$, $k$
    \Output Proof P
    \State $\mathbf{r}_0 \leftarrow \mathbf{0} $ \Comment{$\mathbf{0}$ is a vector of 0s}
    \State P = \Call {Prove}{$\goal$, $\mathbf{r}_0$, []}
    \Function {Prove}{Q, $\mathbf{r_t}$, stack}
        \State embed $Q$ using the character RNN to obtain $\mathbf{q}_t$
        \State input $\mathbf{q}_t$ and $\mathbf{r}_t$ to the model and compute $\mathbf{r}_{t+1}$ (Equation~\eqref{eq:ct})
        \State compute $c_t$ (Equation~\eqref{eq:ct})
        \State $\{ R^{1} \dots R^{k} \} \leftarrow $ From $M^{rule}$ retrieve $k$ entries corresponding to the top $k$ entries of $\mathbf{r}_{t+1}$ 
        \For{$i \in [0,k]$} 
            \State $SU \leftarrow$ \Call {Soft\_Unify} {$Q$, head($R^i$)}
            \If {$SU == False$}
                \State continue to $i+1$
            \Else
                \If {$c_t > T_2$}
                	\State \Return stack
                \EndIf
                \State add $R^{i}$ to stack
                \State \Call {Prove}{Body($R^i$), $\mathbf{r}_{t+1}$, stack} \Comment{Prove the body of $R^i$}
            \EndIf
        \EndFor
        \Return stack
    \EndFunction
    \Function {Soft\_Unify} {$G$, $H$}
        \If {arity($G$) $\neq$ arity($H$)}
            \State \Return False
        \EndIf
        \State Use $\mathbf{M}^{var}$ to compute cosine similarity $S_i$ for all corresponding variable pairs in $G$ and $H$
        \If {$S_i > T_1 \quad \forall \quad i \in [0,\text{arity}(G)]$}
            \State \Return True
        \Else
            \State \Return False
        \EndIf
    \EndFunction
\end{algorithmic}
\end{algorithm}

\subsection*{Extended Discussion}
Table \ref{tab:user_study_lt} shows the performance breakdown with respect to the logic templates in Table \ref{tab:logic_templates}. 
Currently, CORGI uses a general theorem prover that can prove all the templates. The large variation in performance indicates that taking into account the different templates would improve the performance. For example, the low performance on the green template is expected, since CORGI currently does not support the extraction of a hidden \textAction from the user, and interactions only support extraction of missing \textGoal s. This interesting observation indicates that, even within the same benchmark, we might need to develop several reasoning strategies to solve reasoning problems. Therefore, even if CORGI adapts a general theorem prover, accounting for logic templates in the conversational knowledge extraction component would allow it to achieve better performance on other templates.

\begin{table}[ht]
    \caption{Number of successful reasoning tasks vs number of attempts under different scenarios. In CORGI's Oracle unification, soft unification is 100\% accurate. LT stands for Logic Template and LTi refers to template $i$ in Table \ref{tab:logic_templates}.}
    \label{tab:user_study_lt}
    \centering
    \resizebox{0.5\textwidth}{!}{%
    \begin{tabular}{lcccc}
    \toprule
    CORGI & \blueTemplate{LT1} & \orangeTemplate{LT2} & \greenTemplate{LT3} & \redTemplate{LT5} \\ \midrule
Oracle Unification & 24\% & 38\% & 11\% & 0\% \\
\bottomrule
    \end{tabular}
    }
\end{table}

\end{document}